\documentclass[sigconf]{acmart}
\usepackage{graphicx}
\usepackage{subcaption}
\usepackage[linesnumbered,ruled,vlined,noend]{algorithm2e}
\usepackage{multirow}
\usepackage{enumerate}
\usepackage{adjustbox}
\graphicspath{{images/}}

\definecolor{comment-color}{rgb}{0.216, 0.455, 0.424}
\AtBeginDocument{%
  \providecommand\BibTeX{{%
    \normalfont B\kern-0.5em{\scshape i\kern-0.25em b}\kern-0.8em\TeX}}}

\setcopyright{none}
\renewcommand\footnotetextcopyrightpermission[1]{}
\begin{document}
\fancyhead{}
\title{Chat-to-Design: AI Assisted Personalized Fashion Design}

\author{Weiming Zhuang$^{1, 2}$ \quad Chongjie Ye$^{3}$ \quad Ying Xu$^{4}$ \quad Pengzhi Mao$^{5,6}$ \quad Shuai Zhang$^{2}$ }
\affiliation{%
 $^{1}$S-Lab, Nanyang Technological University \enspace $^{2}$SenseTime Research \\ $^{3}$The Chinese University of Hong Kong \enspace $^{4}$Wuhan University of Technology \\ $^{5}$Key Lab of Intell. Info. Process., Inst. of Comput. Tech., CAS 
 \enspace $^{6}$ University of Chinese Academy of Sciences 
}

\renewcommand{\authors}{Weiming Zhuang, Chongjie Ye, Ying Xu, Pengzhi Mao, and Shuai Zhang}

\renewcommand{\shortauthors}{W. Zhuang et al.}

\begin{abstract}

In this demo, we present Chat-to-Design, a new multimodal interaction system for personalized fashion design. Compared to classic systems that recommend apparel based on keywords, Chat-to-Design enables users to design clothes in two steps: 1) coarse-grained selection via conversation and 2) fine-grained editing via an interactive interface. It encompasses three sub-systems to deliver an immersive user experience: A conversation system empowered by natural language understanding to accept users' requests and manages dialogs; A multimodal fashion retrieval system empowered by a large-scale pretrained language-image network to retrieve requested apparel; A fashion design system empowered by emerging generative techniques to edit attributes of retrieved clothes.


\end{abstract}



\maketitle

\section{Introduction}

Searching for desired clothes to buy on e-commerce websites has evolved into a new habit of the new generation. However, classic recommendation systems are difficult to fully capture users' preferences based on clicking or rating data \cite{bell2007lessons}. Besides, ready-made clothes may not satisfy the growing needs of personalization of the new generation. This demo presents a new multimodal interaction system for personalized fashion design, Chat-to-Design. It understands users' intentions via conversation and enables them to further customize clothes. 

\begin{figure}[t]
    \centering
    \begin{subfigure}[h]{0.2\textwidth}
        \includegraphics[width=\textwidth]{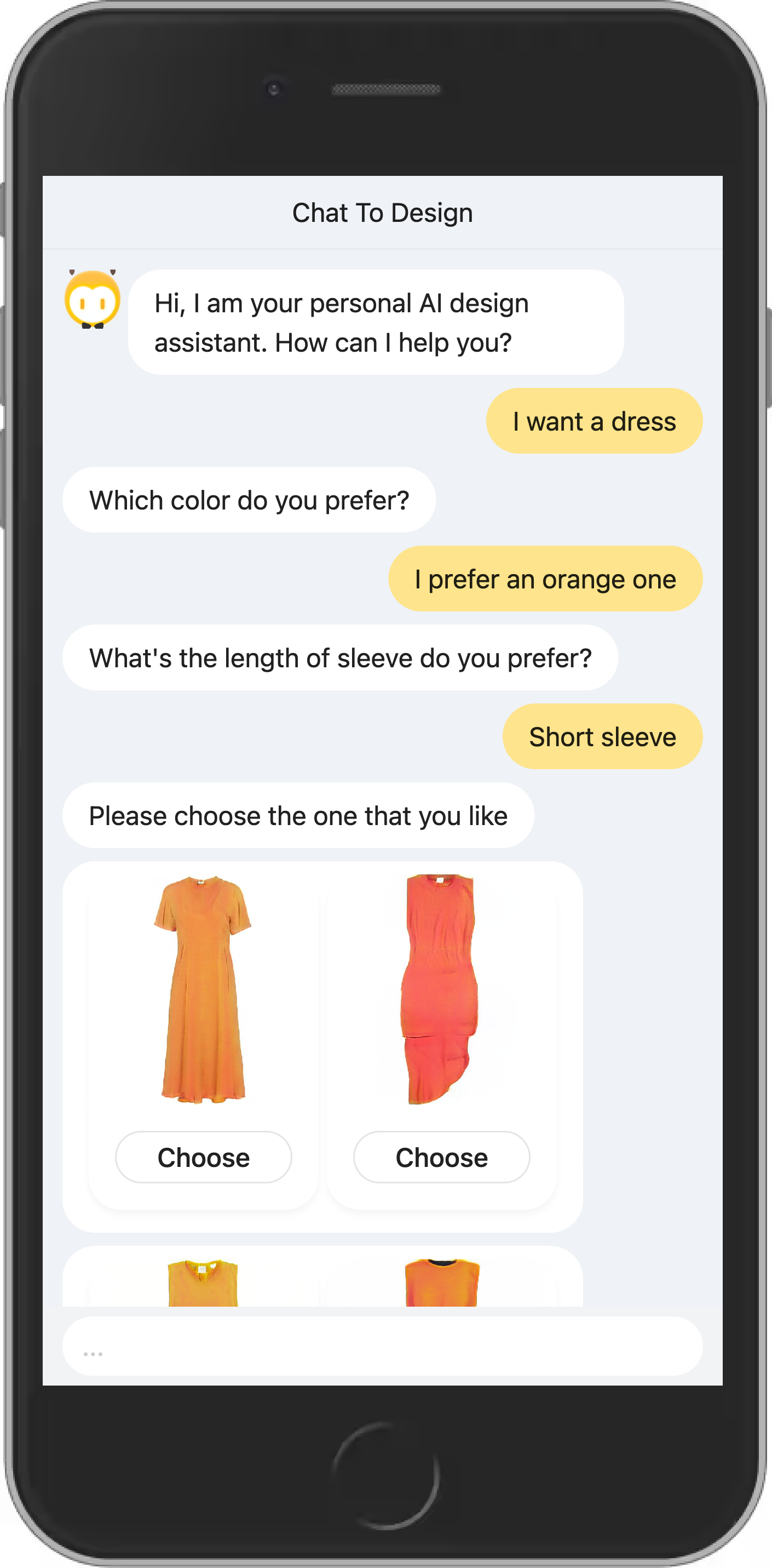}
        \caption{Retrive clothes via dialog}
        \label{fig:ui1}
    \end{subfigure}
    \hspace{1em}
    \begin{subfigure}[h]{0.2\textwidth}
       \includegraphics[width=\textwidth]{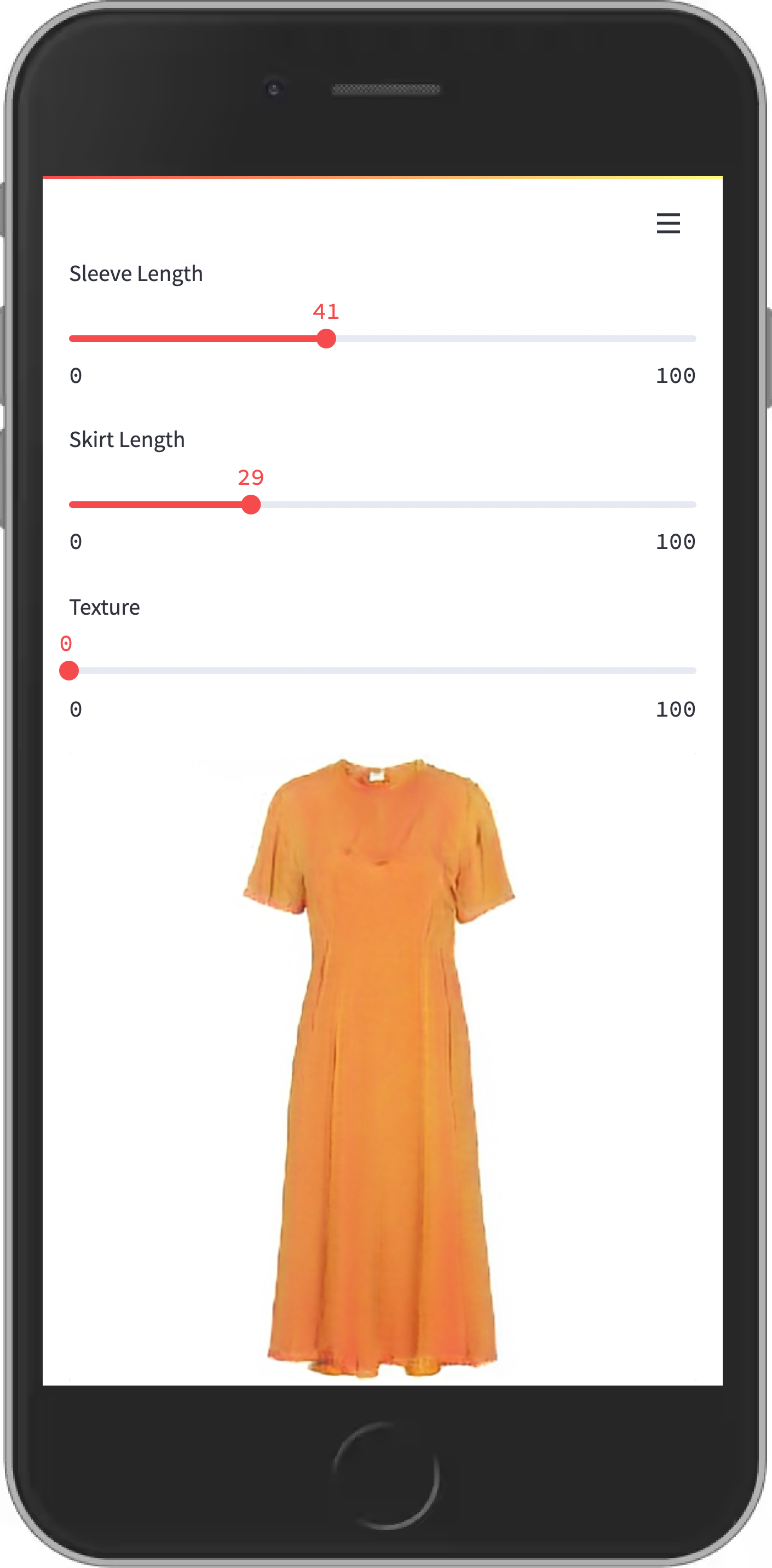}
       \caption{Design detials via editing}
       \label{fig:ui2}
   \end{subfigure}
   \caption{User interfaces and illustrations of Chat-to-Design.}
   \label{fig:ui}
\end{figure}

The user-friendly interface of Chat-to-Design encompasses a complex system that integrates \emph{conversation system} empowered by natural language understanding and dialog system, \emph{multimodal fashion retrieval system} empowered by a large-scale language-image pre-trained network, and \emph{fashion design system} empowered by the domain-specific generative networks. Chat-to-Design helps users design favored cloth in two-step: coarse-grained selection via conversation in Figure \ref{fig:ui1} and fine-grained adjustment via interactive interface in Figure \ref{fig:ui2}. To the best of our knowledge, we are the first to explore such a comprehensive multimodal interaction system for design in the multimedia area.


\begin{figure*}[h]
    \centering
    \includegraphics[width=0.7\textwidth]{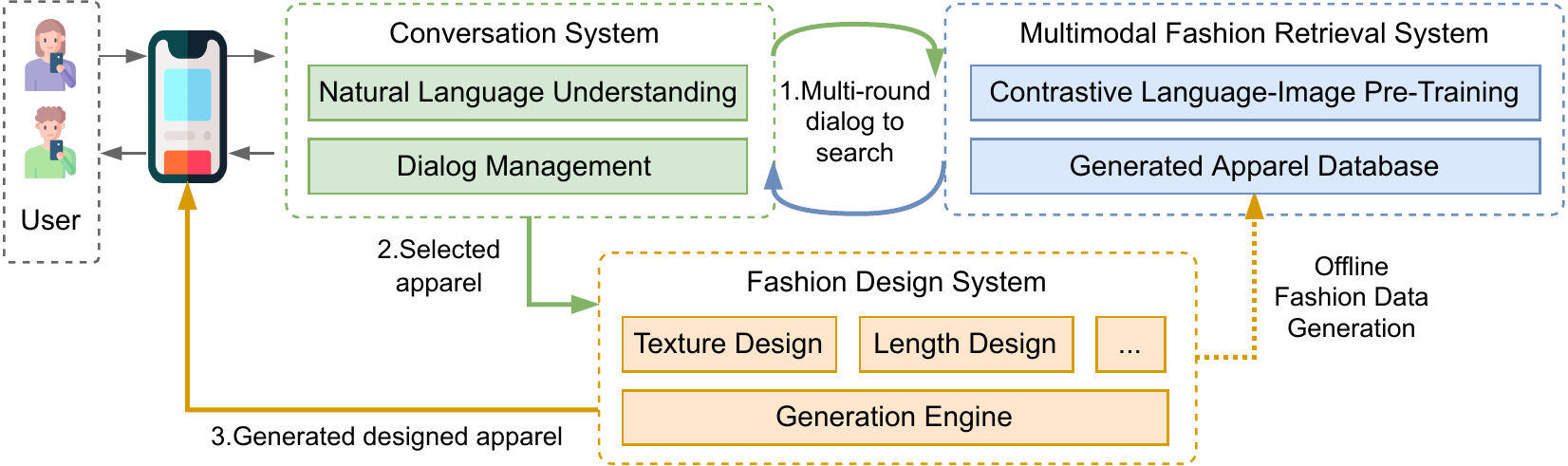}
    \caption{The system architecture of Chat-to-Design. It includes three sub-systems: conversation system, multimodal fashion retrieval system, and fashion design system.}
    \label{fig:architecture}
\end{figure*}

Our system is different from the existing system in many different aspects. Different from traditional keyword-based or filter-based search systems, Chat-to-Design employs dialog to interpret the needs of users. Different from a pure chatbot \cite{wu2021fashioniq} or pure multimodal fashion retrieval system \cite{liao2018interpretable}, Chat-to-Design not only provides a dialog system but also embraces multimodal interactions. Different from existing method on AI-assisted fashion design \cite{rostamzadeh2018fashion,dubey2020ai}, Chat-to-Design builds an end-to-end multimodal interaction system and leverages the lastest generative approach, StyleGAN \cite{karras2019stylegan} StyleFlow \cite{abdal2021styleflow} to enable modifying apparel based on attributes.

\section{System Architecture}

The system is designed to be user-friendly and real-time interactive. Figure \ref{fig:architecture} depicts the system architecture of Chat-to-Design. It comprises three sub-systems: \emph{conversation system} accepts and interprets users' requests, manages the dialog, and sends requests to the other two systems; \emph{multimodal fashion retrieval system} retrieves the most appropriate apparel according to description; \emph{fashion design system} enables users to further edit the details of the selected apparel. Next, we introduce these three systems in detail.

\subsection{Conversation System}

The conversation system is the entrance of the whole system. It consists of a natural language understanding (NLU) module and a dialog management module. NLU has been a well-studied research field. We adopt the widely-used MitieNLP \cite{dlib09mitinlp} toolkit for intent recognition and entity extraction inside the NLU module. To equip the NLU module with domain knowledge of fashion, we generate synthesized fashion dialog data using Chatito \cite{chatito}. The dialog management manages multi-round conversations. We integrate these two modules using the widely-adopted Rasa \cite{rasa} framework. The conversation system responds with multimodal contents: text, images, and actions, which greatly enriches users' experience.

\subsection{Multimodal Fashion Retrieval System}

Multimodal fashion retrieval system searches and returns suitable clothes with given extracted user requests. Multimodal search is attracting considerable attention. We build this system based on CLIP (Contrastive Language-Image Pre-Training) \cite{radford2021clip}, which is pretrained on 400 million pairs of images and text pairs from the Internet. CLIP is powerful in establishing the connection between image and text. We further empower it with domain knowledge of fashion by fine-tuning its pretrained Vision Transformer \cite{dosovitskiy2021vit} using Fashion200K dataset \cite{han2017automatic}, which contains 200K image and text pairs.

\subsection{Fashion Design System}

\begin{figure}[t]
    \centering
    \begin{subfigure}[h]{0.4\textwidth}
        \includegraphics[width=\textwidth]{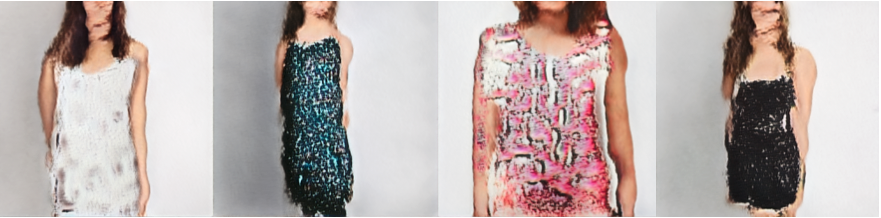}
        \caption{Clothes generated using existing datasets}
        \label{fig:gen1}
    \end{subfigure}
    \begin{subfigure}[h]{0.4\textwidth}
       \includegraphics[width=\textwidth]{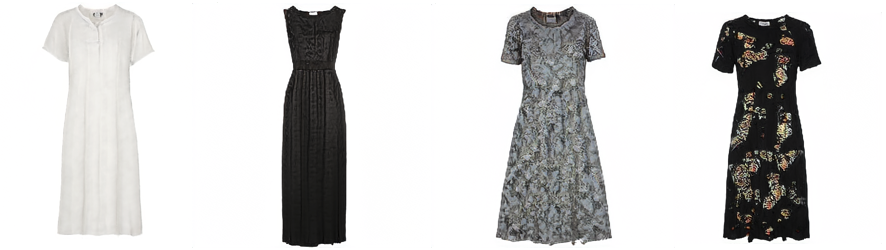}
       \caption{Clothes generated using our newly constructed dataset}
       \label{fig:gen2}
   \end{subfigure}
   \caption{Generated cloth samples using different datasets.}
   \label{fig:comparison}
\end{figure}

We support real-time fashion attribute editing in the fashion design system. It allows users to flexibly design clothes by adjusting attributes like texture and sleeve length. The system is empowered by the latest generative techniques, including StyleGAN \cite{karras2019stylegan} and StyleFlow \cite{abdal2021styleflow}. To the best of our knowledge, we are the first to adopt both StyleGAN and StyleFlow to the fashion domain. We train these models with a newly constructed large-scale fashion dataset to enable them with knowledge in the fashion domain. In particular, we construct the dataset by first collecting around 100,000 images of 6 attributes and merging it with Fashion200K \cite{han2017automatic} data. Then, we clean the dataset by removing the background and human bodies with semantic segmentation and edge detection techniques. Figure \ref{fig:comparison} illustrates that clothes generated using our clean dataset are superior to using existing datasets like Fashion200K \cite{han2017automatic}. Besides, we generate fashion data offline and merge it into the database for multimodal fashion retrieval.

\section{Demonstration}

Figure \ref{fig:ui} demonstrates the user interfaces and workflow of interacting with the system. Chat-to-Design helps design users favored cloth in two steps: coarse-grained selection via conversation in Figure \ref{fig:ui1} and fine-grained adjustment via interactive interface in Figure \ref{fig:ui2}. Firstly, the user starts a conversation with a query sentence like "I want a dress". The system chats with the user to collect other desired attributes of the dress, such as color and sleeve length. After that, the system suggests multiple suitable dresses, which the user can either select the desired dress or provide further description. After selecting a dress, the user can further tune the details of attributes of clothes, such as adjusting the textures and changing sleeve length. Chat-to-Design enables the user to view the effect of these changes immediately and confirm the desired design.

Chat-to-Design has a broad scope of potential applications. It can be integrated into e-commerce websites to better capture consumers' needs and support the customer-driven manufacturing process. Besides, by further integrating virtual try-it-on, users can share their photos wearing self-designed clothes on social media and short videos. Lastly, Chat-to-Design has the potential to support user-designed virtual apparel in the metaverse.  

\begin{acks}
    This study is supported by the RIE2020 Industry Alignment Fund - Industry Collaboration Projects (IAF-ICP) Funding Initiative, as well as cash and in-kind contribution from the industry partner(s). We would also like to thank DeeCamp organizing committee for providing computing resources and generous support.
\end{acks}

\bibliographystyle{ACM-Reference-Format}
\bibliography{references}

\newpage

\end{document}